\documentclass[a4paper,fleqn]{cas-dc}

\usepackage[numbers]{natbib}
\usepackage{epstopdf}
\usepackage{color}

\usepackage{booktabs}
\usepackage{subfigure}
\usepackage{multirow}
\usepackage{amsmath,amsfonts}
\usepackage[ruled,linesnumbered]{algorithm2e}

\def\tsc#1{\csdef{#1}{\textsc{\lowercase{#1}}\xspace}}
\tsc{WGM}
\tsc{QE}
\tsc{EP}
\tsc{PMS}
\tsc{BEC}
\tsc{DE}
\begin{document}
\let\WriteBookmarks\relax
\def\floatpagepagefraction{1}
\def\textpagefraction{.001}
\shorttitle{Improving Fraud Detection via Hierarchical Attention-based Graph Neural Network}
% \shortauthors{L. Li et~al.} %% 缩略作者 自己名字， 比如： 张三 = S. Zhang

%% 标题
\title [mode = title]{Improving Fraud Detection via Hierarchical Attention-based Graph Neural Network}                      
%%\tnotemark[1,2]

%%\tnotetext[1]{This document is the results of the research project funded by the National Science Foundation.}

%%\tnotetext[2]{The second title footnote which is a longer text matter to fill through the whole text width and overflow into another line in the footnotes area of the first page.}

%% 作者顺序
%% 1
\author{Yajing Liu}
% \ead{liuyajing2019@ia.ac.cn}

\address{Research Center of Precision Sensing and Control, Institute of Automation, Chinese Academy of Sciences, Beijing, China}

\author{Zhengya Sun}
\cormark[1]
\ead{zhengya.sun@ia.ac.cn}

\author{Wensheng Zhang}
% \ead{zhangwenshengia@hotmail.com}

\address{University of Chinese Academy of Sciences, Beijing, China}

\cortext[cor1]{Corresponding author:} %% 首页左下角通讯作者
%%\cortext[cor2]{Principal corresponding author} 

%%\fntext[fn1]{This is the first author footnote. but is common to thirdauthor as well.}
%%\fntext[fn2]{Another author footnote, this is a very long footnote and it should be a really long footnote. But this footnote is not yet sufficiently long enough to make two lines of footnote text.}

%%\nonumnote{This note has no numbers. In this work we demonstrate $a_b$ the formation Y\_1 of a new type of polariton on the interface between a cuprous oxide slab and a polystyrene micro-sphere placed on the slab.}

%%摘要
\begin{abstract}
Graph neural networks (GNN) have emerged as a powerful tool for fraud detection tasks, where fraudulent nodes are identified by aggregating neighbor information via different relations. To get around such detection, crafty fraudsters resort to camouflage via connecting to legitimate users (i.e., relation camouflage) or providing seemingly legitimate feedbacks (i.e., feature camouflage). A wide-spread solution reinforces the GNN aggregation process with neighbor selectors according to original node features. This method may carry limitations when identifying fraudsters not only with the relation camouflage, but with the feature camouflage making them hard to distinguish from their legitimate neighbors. In this paper, we propose a Hierarchical Attention-based Graph Neural Network (HA-GNN) for fraud detection, which incorporates weighted adjacency matrices across different relations against camouflage. This is motivated in the Relational Density Theory and is exploited for forming a hierarchical attention-based graph neural network. Specifically, we design a relation attention module to reflect the tie strength between two nodes, while a neighborhood attention module to capture the long-range structural affinity associated with the graph. We generate node embeddings by aggregating information from local/long-range structures and original node features. Experiments on three real-world datasets demonstrate the effectiveness of our model over the state-of-the-arts.
\end{abstract}

% %%\includegraphics{figs/grabs.pdf} %%图片摘要地址路径

%%高亮
% \begin{highlights}
% \item highlights 1.
% \item highlights 2.
% \item highlights 3.
% \end{highlights}

%% 关键词

% 此指令为生成标题格式，不可删除
\maketitle  

%% 1.引言
\section{Introduction}

With the rapid development of the Internet, social media has become a universal platform for users to communicate and obtain information. However, many fraudsters appear in social media \cite{appear}. These fraudsters send spam messages \cite{OSSD}, post fake reviews \cite{FdGars}, and commit financial fraud \cite{HACUD}, seriously interfering with normal online activities. In the face of potential social panic and financial threats, detecting fraud on social media is critical.

One straightforward way for fraud detection is to leverage a heterogeneous graph to capture multiple types of relations among different entities (e.g., users,  reviews, and products that the users have reviewed), and build supervised classifiers with the features extracted based on the graph properties, \cite{graphBased1, graphBased2}. These approaches require considerable domain knowledge to develop good features, thus time-consuming and labor-intensive. Moreover, those handcrafted features fail to exploit high-level representations in the correlation between entities.

Recent progress has been made on graph neural networks on heterogeneous graphs. Due to the ability to automatically learn more comprehensive representations beyond human-designed ones, an increasing number of approaches attempt to develop GNN-based fraud detection frameworks. This line of research is mainly focused on developing aggregation methods from different perspectives. 
For instance, some approaches only mine the correlation features within local neighbors by employing convolution layers \cite{FdGars,GAS,GEM}. While other methods aggregate information intra-view and inter-view hierarchically, which convolute each node over its within-view neighborhood and then integrate information across views \cite{GeniePath,SemiGNN,Player2Vec,HACUD}.
A common characteristic is that they extract the clustering behaviors of fraudsters through aggregating neighborhood information.
Yet, a major limitation is that neighbor aggregation, though fewer data annotation cost, may be confused by camouflaged fraudsters. In reality, smart fraudsters usually gloss over explicit suspicious by feature camouflage and relation camouflage \cite{CARE-GNN}. If we aggregate neighbors with the legitimate reviews as node features, it will probably eliminate the suspiciousness of the center fraudster.

Instead of using all adjacent neighbors, some researchers proposed to aggregate only a small subset of the neighbors for againsting camouflaged fraudsters. Obviously, the key issue is how to select this neighbor subset. A good candidate should allow the non-selected neighbors to be dissimilar or inconsistent to a center node within a relation \cite{CARE-GNN,GraphConsis}. However, it is often the case that the center nodes are fraudsters not only with the relation camouflage, but with the feature camouflage making them hard to distinguish from their legitimate neighbors. In a sense, these approaches cannot deal well with the suspiciousness of center nodes that own almost identical features with legitimate ones, but rather downplay the difficulty by assuming them with non-disguised features. 

In fact, there are fraudsters not only with the relation camouflage, but also with the feature camouflage. Consider the online shopping platform scenario, where a group of fraudsters engage in posting inauthentic content and unfair ratings to promote (or defame) a target product. To bypass fraud detectors, besides connecting themselves to regular entities (i.e., relation camouflage), they provide feedbacks that look legitimate, and only once in a while inject fake feedbacks (i.e., feature camouflage). As demonstrated by a synthetic example, the center node is a fraudster who connects to a legitimate user (i.e., reviewing the same product within one week), and three fraudsters (i.e., reviewing the same product, giving the same rating within one week, posting reviews with similar content within one week.). Among three fraudsters, only one fraudster always provides fake feedbacks, which would probably be filtered under each relation in view of neighbor selectors. Apparently, aggregating features from those more similar neighbors severely hinders the recognition of the center fraudster. As a matter of fact, it is often the case that fraudsters are more densely connected than legitimate users, regardless of whether fraudsters camouflage themselves \cite{similcatch}.

\begin{figure*}
\centerline{\includegraphics[width=0.75\textwidth]{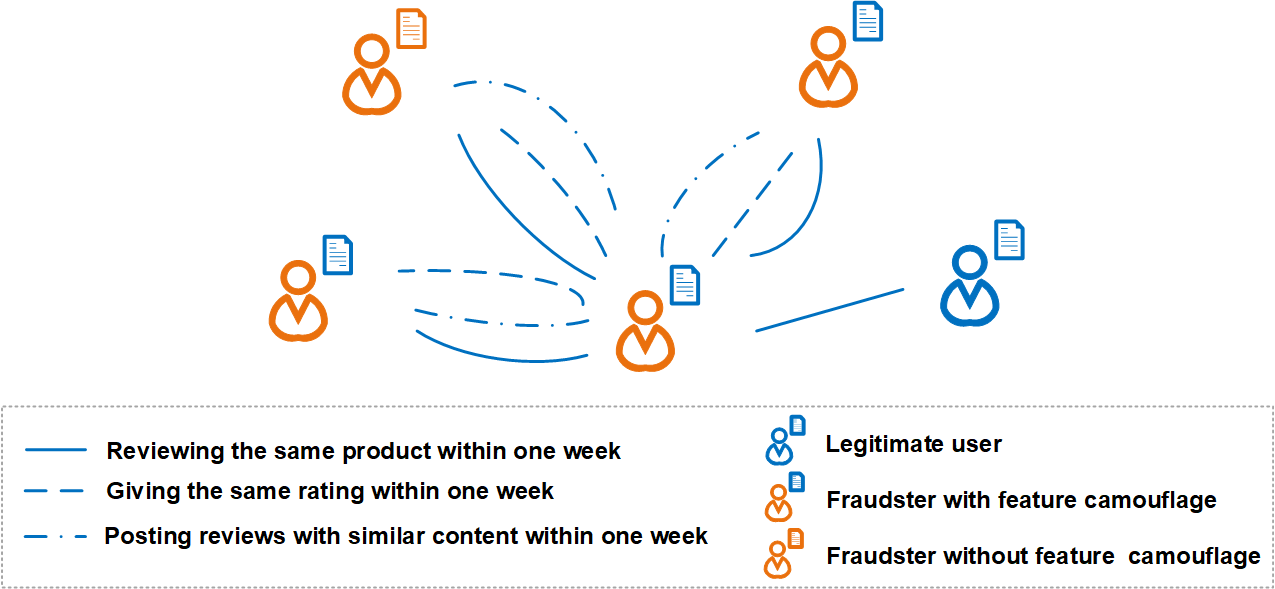}}
\caption{A situation of a camouflaged fraudster and its neighbors. Suppose we encode the center fraudster by aggregating features of all neighbors similar to traditional GNN models. In that case, features from legitimate neighbors and camouflaged fraudster neighbors will probably smooth out the suspicion of the center fraudster. Suppose we target most similar neighbors (i.e., camouflaged fraudsters and legitimate users) similar to filter-based GNN methods. In that case, the suspicion of the center fraudster will be eliminated.}
\label{fig-example}
\end{figure*}

In this paper, we propose a novel Hierarchical Attention-based GNN model (HA-GNN) for fraud detection, which integrates the local and long-range structural encoding into original node features for better representation learning. HA-GNN introduces a relation attention module, and exploits the adjacency matrices across different relations for mining the local structural information. This is helpful for characterizing the tie strength between every two nodes. HA-GNN then exploits a multi-head self-attention mechanism learns to weigh the neighbors depending on their structural affinity when aggregating information from the neighbors. After that, HA-GNN merges the complementary information of structures and original features so as to achieve more informative representations. The experimental results on three real-world datasets validate the efficacy of our proposed model compared to the state-of-the-arts.

The main contributions of this work are as follows: 
\begin{itemize}
\item  We offer a novel hierarchical attention-based GNN model, which encodes both structures and original features by introducing a 3-level attention mechanism, thus augments the GNN model's capability to identify fraudsters.

\item  We grasp the structural information in both local and long-range scopes. This is implemented by designing two attention modules, i.e., relation attention module and neighborhood attention module. The former characterizes the tie strength between two adjacent nodes, while the latter captures the structural affinity associated with the graph.

\item  We conduct extensive experiments to demonstrate the effectiveness of the proposed HA-GNN in recognizing fraudsters. Our model empowered by three attention modules achieves superior performance on three real-world social network datasets.
\end{itemize}

\section{RELATED WORK}
\subsection{Graph Neural Network (GNN)}

GNNs have made prominent progress in graph representation learning. Motivated by Convolutional Neural Networks (CNNs) \cite{Alexnet,googlenet,resnet,cnnsurvey}, GNNs generalize the convolution operator from grid data to graph data \cite{survey_gnn}. They generate the representation of the target node by aggregating neighbors’ and itself features. The key difference lies in the way messages are propagated. GCN \cite{GCN} propagates information based on a graph Laplacian matrix and assigns nonparametric weights to neighbors during the aggregation process. GraphSAGE \cite{GraphSAGE} adopts a sampling mechanism to obtain a fixed number of neighbors for each node. GIN \cite{GIN} employs the sum-like aggregation function and adjusts the weight of the central node by a learnable parameter. While these GNNs assume that the influence of neighboring nodes to the central node is equal, GAT \cite{GAT} adopts an attention mechanism to learn the relative weights between two connected nodes. Gated Attention Network (GAAN) \cite{GAAN} performs a novel
multi-head attention mechanism that learns an additional attention weight for each attention head. Apart from applying graph attention spatially, GeniePath \cite{GeniePath} introduces an LSTM-like gating operator to control messages passing across graph convolutional layers. These models have been widely used in various real-world applications, such as recommendation systems \cite{Recommender_1,Recommender_2,Recommender_3}, traffic forecasting \cite{Traffic_1,Traffic_2,Traffic_3} and fraud detection \cite{PCGNN,RAREGNN,IHGAT}. This paper employs the attention mechanism for fraud detection with camouflage.

\subsection{Graph Fraud Detection}

Graph algorithms have long been considered as important tools in fraud detection. Early researchers \cite{graphBased1,graphBased2} established graph analysis techniques for fraud detection by extracting graph-centric features, measuring the closeness of nodes, and finding densely connected groups in the graph. Nevertheless, these attempts usually rely on human-defined rules or features, which are not easy to generalize to various datasets. In recent years, graph neural networks have gained significant interest. Their applications in fraud detection have yielded promising results. HACUD \cite{HACUD} identifies cash-out users in the scenario of credit payment service. Player2Vec \cite{Player2Vec} detects key players in underground forums. GAS \cite{GAS} identifies spam advertisements at Xianyu. FdGars \cite{FdGars} distinguishes fraudsters from normal users in the Tencent App Store. SemiGNN \cite{SemiGNN} discovers fraudsters in financial networks. These models extend graph convolutional networks, graph attention networks, and hierarchical graph neural networks to aggregate neighborhood information, enhancing feature representations of objects/users. However, they neglect the inconsistency problem between nodes and their neighbors caused by  fraudsters camouflage.

One way to tackle the inconsistency problem is to focus on similar neighbors. GraphConsis \cite{GraphConsis} samples neighbors through a consistency score. CARE-GNN \cite{CARE-GNN} proposes a label-aware aggregator and a similarity-aware neighbor selector to filter inconsistent neighbors. PC-GNN \cite{PCGNN} devises a label-balanced sampler and a neighborhood sampler to select nodes and their similar neighbors. They generate node representations by aggregating information from selected neighbors. Another way to alleviate the inconsistency problem is decoupling representation learning and classification. DCI \cite{DCI} and GCCAD \cite{GCCAD} develop a self-supervised graph learning scheme to learn comprehensive representations. PAMFUL \cite{PAMFUL} uses a GNN encoder to perform feature aggregation and pattern mining algorithms to supervise the GNN training process. These works show limitations in identifying fraudsters that have both feature camouflage and relation camouflage. To improve the capability of fraud detection, our approach exploits adjacency matrices across multiple relations to capture tie strength between each two nodes, and merges information of structures and original node features via the attention mechanism.

\begin{figure*}
\centering
\includegraphics[width=1\textwidth]{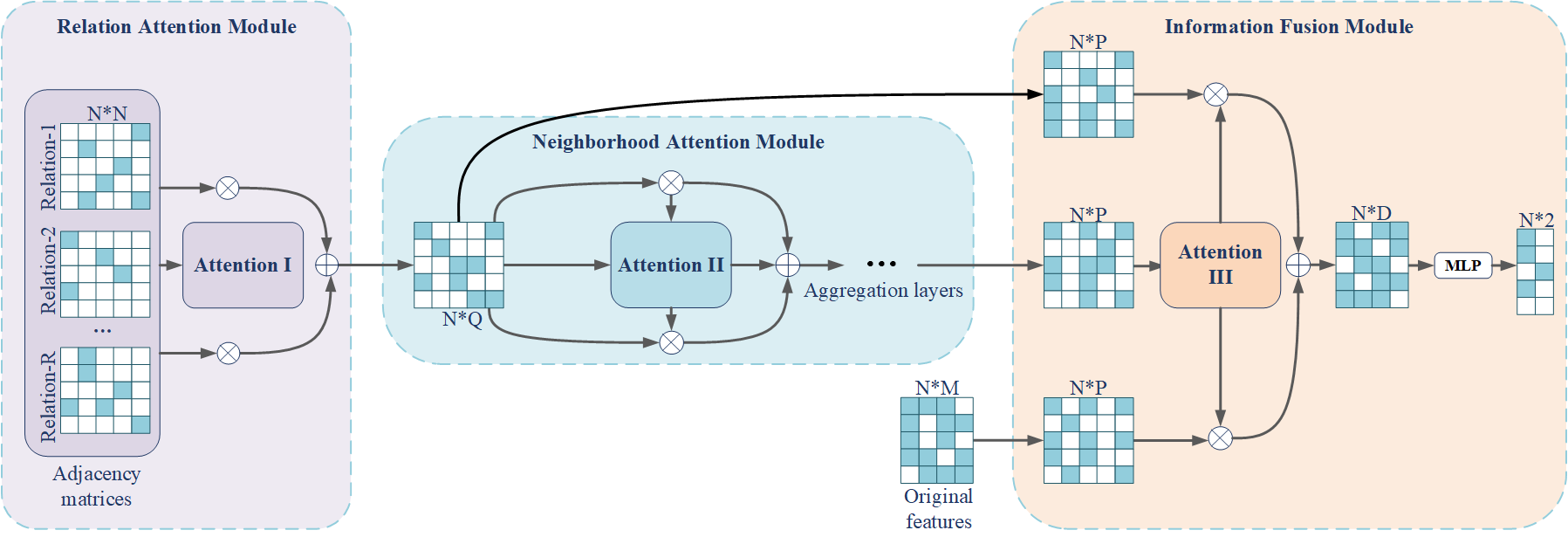}
\caption{\centering{The framework of the proposed model.}}
\label{fig-model}
\end{figure*}        

\section{THE PROPOSED MODEL}
\subsection{Model Overview}
We now describe HA-GNN, a hierarchical attention-based GNN model for fraud detection. HA-GNN is composed of three hierarchical processes to deal with fraudsters: 1) encoding each node locally according to adjacency matrices across different relations, 2) aggregating neighbors with multi-head self-attention for multiple layers to convey long-range structural characteristics, and 3) generating node embeddings by jointly modeling structure (local and long-range) and feature information. The pseudocode of the overall algorithm is provided in Algorithm \eqref{alg_1}.

\subsection{Notation}
We define a heterogeneous graph (i.e., multi-relation graph) as $\mathcal{G}=\left \{\mathcal{V}, \mathcal{X}, \{\mathcal{E}^r\}|_{r=1}^R, \{ \mathcal{A}^r\}|_{r=1}^R, \mathcal{Y} \right \}$, where $\mathcal{V} = \left\{v_1 ,..., v_{N}\right\}$ denotes the node set, $\mathcal{Y}$ denotes the set of labels for each node in $\mathcal{V}$, $\mathcal{X} = \left\{f_{1}, ..., f_{N}\right\}$ denotes the original feature 
set and $f_{i}$ denotes original feature of node $i$. $\mathcal{E}^r$ denotes the edges w.r.t. relation $r$. Note that an edge can be associated with multiple relations and there are $R$ different types of relations. $\mathcal{A}^r$ denotes adjacency matrix of sub-graph corresponding to relation $r$. Given $\mathcal{G}$, the goal of HA-GNN is to learn a semi-supervised binary node classifier.

\subsection{Relation Attention Module}

Relational Density Theory (RDT) \cite{RDT} shows that relational mass is associated with relational density and relational volume. Inspired by the theory, we incorporate weighted adjacency matrices across multiple relations to measure tie strength between two nodes.

Specifically, with $\{ \mathcal{A}^r\}|_{r=1}^R = \{a^{r}_1, ..., a^{r}_{N} \}|_{r=1}^R$ as input, we calculate the importance of relation $r$ through a relation-level function as follows:

\begin{equation}
    w_{r}=\dfrac{1}{n}\sum^{N} _{i=1}q^\top\tanh\left(W\cdot a^{r}_i + b\right)\label{eq_w}
\end{equation}
where $W$ and $b$ denote the weight and bias of the nonlinear transformation, respectively. $q$ denotes an attention vector. Note that these three parameters are shared for $\{\mathcal{A}^r\}|_{r=1}^R$.

After obtaining the importance of each relation, we normalize them through a softmax function to calculate the weight of relation $r$ as follows:

\begin{equation}
    \beta _{r}=\dfrac{\exp \left( w_{r} \right) }{\sum ^{R}_{r=1}\exp \left( w_{r} \right) }\label{eq_beta}
\end{equation}

With the learned weights, we calculate the weighted sum of $\{a^r_i\}|^R_{r=1}$ to learn the local embedding (i.e., the output of the relation attention module) as follows:

\begin{equation}
    h_{i} =\sum ^{R}_{r=1}\beta _{r}\cdot a^{r}_i \label{eq_h_v}
\end{equation}

\subsection{Neighborhood Attention Module}

Given local embeddings $\{h_{i}, \forall i\in \mathcal{V}\}$ generated from the relation attention module, we aggregate neighbors from all relations. Moreover, we leverage the multi-head self-attention mechanism \cite{self-attention} to calculate the importance of each neighbor. 
We first obtain weight $\alpha^{\left(l\right)}_{ij}$ of node $j \rightarrow i$ at the $l$-th layer ($l \in\{1,...,L\}$). Specifically, we concatenate $g^{\left(l-1\right)}_{i}$ and $g^{\left(l-1\right)}_{j}$, where $g^{\left(l-1\right)}_{i}$ is the embedding of node $i$ at the $\left(l-1\right)$th layer and $g^{\left(0\right)}_{i} = h_{i}$. Then we compute dot products of the concatenation and a weight vector, and apply the activation function and the softmax function on it. Note that the weight is asymmetric, which implies $\alpha^{\left(l\right)}_{ij}$ and $\alpha^{\left(l\right)}_{ji}$ are not equal. We calculate the weight $\alpha_{ij}$ at the $l$-layer as follows:

\begin{equation}
\begin{split}
\alpha^{\left(l\right)}_{ij} = \frac{exp \left( \sigma \left( a^{\left(l\right)\top} \cdot \left[g^{\left(l-1\right)}_{i}\|g^{\left(l-1\right)}_{j} \right] \right)\right )}{\sum _{j \epsilon \mathcal{N}_{i}}exp \left(\sigma\left(a^{\left(l\right)\top} \cdot \left [g^{\left(l-1\right)}_{i}\|g^{\left(l-1\right)}_{j} \right] \right)\right)} \label{eq_alpha_uv}
\end{split}
\end{equation}

Here, $\sigma$ denotes the activation function, and $\|$ denotes the concatenate operator. $\mathcal{N}_{i}$ denotes neighbors of node $i$ from all relations (including node $i$ itself), i.e., $\mathcal{N}_{i} = \{j, \forall \left(i, j\right) \in \mathcal{E}^r|^R_{r=1}\}\cup\{i\}$. Furthermore, $a^{\left(l\right)}$ denotes the weight vector at the $l$-layer. The similarity between concatenation $\left [g^{l-1}_{i}\|g^{l-1}_{j} \right ]$ and vector $a^{l}$ determines the weight of node $j$ to node $i$. 
After that, we generate the hidden embedding of node $i$ as follows:

\begin{equation}
    g^{\left(l\right)}_{i} =\sigma \left(\sum _{j\in \mathcal{N}_{i}}\alpha^{\left(l\right)}_{ij}\cdot g^{\left(l-1\right)}_{j}\right) \label{eq_z_v}
\end{equation}

In addition, we employ multi-head attention mechanism \cite{self-attention} to jointly attend to information from different representation subspaces. Specifically speaking, we repeat the process from Algorithm \eqref{alg_1} line \eqref{line_8} to line \eqref{line_10} for $K$ times and then calculate the output vector at the $l$-th layer as follows:

\begin{equation}
    g^{\left(l\right)}_{i}=\left|\right|^{K}_{k=1}\sigma \left(\sum _{j\in \mathcal{N}_{i}}\alpha^{k, \left(l-1\right)}_{ij}\cdot g^{\left(l-1\right)}_{j}\right)\label{eq_z_v_k}
\end{equation}

For each node $i$, the long-range embedding $g_{i}$ is the output of the $L$-th layer in the neighborhood attention module, i.e., $g_{i} = g^{\left(L\right)}_{i}$.

\subsection{Information Fusion Module}

We generate final node embeddings by aggregating original node features and the output embeddings of the first two modules. Besides, we employ information-level attention \cite{HAN} to obtain weights associated with them. 

Original node feature $f_{i}$ and out embeddings obtained from the first two modules (i.e., $h_{i}$ and $g_{i}$) may be in different feature spaces. We first project them into the same feature space through a multi-layer perceptron\cite{MLP}:

\begin{equation}
 M\left(e_{i} \right) = MLP\left(e_{i}\right) \label{eq_M'}
\end{equation}
where $e_{i}\in\{ h_{i}, g_{i}, f_{i} \}$, $MLP$ denotes the multi-layer perceptron.

After that, we feed projected vectors into a nonlinear transformation. Then, we calculate their importance through the information-level attention function as follows:

\begin{equation}
 \eta\left(e_{i} \right) = p^\top\tanh\left(W'\cdot M\left(e_{i} \right) + b'\right) \label{eq_phi'}
\end{equation}

Here, $W'$ denotes the weight matrix, $b'$ denotes the bias vector, and $p$ denotes the attention vector. Note that for a fair comparison, all of the above parameters are shared for $\{ h_{i}, g_{i}, f_{i}\}$. After obtaining information-level attention function $\{\eta\left(h_{i} \right), \eta\left(g_{i} \right), \eta\left(f_{i} \right)\}$, we normalize them via a softmax function. The information weight of $e_{i}$, denoted as $\phi\left(e_{i} \right)$, can be calculated as follows:

\begin{equation}
 \phi\left(e_{i} \right) =\dfrac{\exp \left(\eta\left(e_{i} \right)\right) }{\sum _{e'_{i}\in\{ h_{i}, g_{i}, f_{i} \}}\exp \left(\eta\left(e'_{i} \right) \right) }\label{eq_phi}
\end{equation}

Intuitively, the larger $\phi\left(e_{i} \right)$ is, the more important $e_{i}$ is. Please note that for different datasets, the weights $\phi\left(e_{i} \right)$ may be difference.
Hence, we calculate the final embedding of node $i$ as follows:

\begin{equation}
 z_{i} = \phi\left(h_{i} \right)\cdot M\left(h_{i}\right) + \phi\left(g_{i} \right)\cdot M\left(g_{i} \right) + \phi\left(f_{i} \right)\cdot M\left(f_{i} \right)\label{eq_final_emb}
\end{equation}

The information fusion module outputs a low-dimensional embedding for each node. These low-dimensional embeddings are fed into a node classifier (implemented with a multi-layer perceptron) to calculate the probability of a node being classified as fraud. 

Legitimate instances in real-world social networks are far more numerous than fraudulent instances. If the loss function ignores the imbalance problem, the neural network will be inclined to overfit to the minority of fraudulent instances. We alleviate the imbalance issue by incorporating the class-balanced cross-entropy loss function \cite{class-balanced-loss}. In particular, we leverage a weight parameter to enhance the weight of labeled data in the minority but more important class. In addition, we reform the semi-supervised classification problem by minimizing the following loss function:

\begin{equation}
\begin{split}
    Loss =-\lambda\sum _{i \epsilon \mathcal{V}_{l_0}}y_{i}\ln\left(\sigma\left(MLP \left(z_{i}\right)\right)\right) \\
    -\sum _{i \epsilon \mathcal{V}_{l_1}}y_{i}\ln\left(\sigma\left(MLP\left(z_{i} \right)\right)\right)\label{eq_loss_info}
\end{split}
\end{equation}
where $\mathcal{V}_{l_1}$ denotes labeled fraudulent instances in the training set, and $\mathcal{V}_{l_0}$ denotes labeled legitimate instances in the training set. $y_{i}$ and $z_{i}$ denote the label vector and final embedding vector of node $i$. $\sigma$ denotes the non-linear activation function (we use the $sigmoid$ function in our work), $MLP$ denotes the multi-layer perceptron, and $\lambda$ denotes the weight parameter to balance the effect of labeled fraudulent instances and labeled legitimate instances. With the guidance of labeled data, we can optimize the proposed model via back propagation and 
obtain the classification of nodes. 

\begin{algorithm} 
  \SetKwInOut{Input}{input}\SetKwInOut{Output}{output}
\caption{The training process of HA-GNN}
\label{alg_1}
\Input{
An undirected heterogeneous graph: $\mathcal{G}=\left \{\mathcal{V}, \mathcal{X}, \{\mathcal{E}^r\}|_{r=1}^R, \{ \mathcal{A}^r\}|_{r=1}^R, \mathcal{Y} \right \}$;
Number of layers, heads: $L$, $K$;
}
\Output{
    The vector representations $\{z_{i}, \forall i\in \mathcal{V}$\}; 
    The weight $\beta$ of each relation; 
    The weight $\alpha$ of each neighbor;
    The weight $\phi$ of each kind of information.
    }
\BlankLine
\For{$i \in \mathcal{V}$}{
    \For{$r=1, \ldots, R$}{
        Calculate relation-specific importance $w_{r}$ by Eq. (\ref{eq_w})\;
        Calculate the weight $\beta^{r}$ by Eq. (\ref{eq_beta})\;
    }
    $h_{i}\leftarrow$ Eq. (\ref{eq_h_v})\;
    $g^{\left(0\right)}_{i}\leftarrow h_{i}$\;
    Obtain neighbors set $\mathcal{N}_{i}$\;
    \For{$l=1, \ldots, L$}{
        \For{$k=1, \ldots, K$}{
            \For{$j \in \mathcal{N}_{i}$}
            {\label{line_8}
                Calculate the weight $\alpha^{k,\left(l\right)}_{ij}$ by Eq.(\ref{eq_alpha_uv})\;
            }\label{line_10}
        }
        $g^{\left(l\right)}_{i}\leftarrow$ Eq. (\ref{eq_z_v_k})\;
    }
    $g_{i}\leftarrow g^{\left(L\right)}_{i}$ \;
    Calculate the weight $\{\phi\left(e_{i}\right), \forall e_{i} \in\{h_i,g_i,f_i\}\}$ by Eq. (\ref{eq_phi})\;
    Calculate the node embedding $z_{i}$ by Eq. (\ref{eq_final_emb})\;
}
Calculate Cross-Entropy $Loss$ by Eq. (\ref{eq_loss_info})\;
Back propagation and update parameters. \\
\Return $z_{i},\forall i\in\mathcal{V}$. 
\end{algorithm}
\DecMargin{1em}

\section{EXPERIMENT}

\subsection{Dataset}

\begin{table*}[width=1\textwidth]
% \begin{table}
\caption{Datasets and graph statistics.}
\begin{center}
\begin{tabular*}{\hsize}{@{}@{\extracolsep{\fill}}llllll@{}}
\toprule
Dataset &
  \begin{tabular}[l]{@{}l@{}}\#Node\\ (Fraud\%)\end{tabular} &
  Relations &
  \#Relations &
  \begin{tabular}[l]{@{}l@{}}Avg.Feature\\ Similarity\end{tabular} &
  \begin{tabular}[l]{@{}l@{}}Avg.Label\\ Similarity\end{tabular} \\ \midrule
\multirow{4}{*}{YelpChi}      & \multirow{4}{*}{\begin{tabular}[c]{@{}l@{}}45,954\\ (14.5\%)\end{tabular}} & $R-U-R$ & 49,315     & 0.83 & 0.90 \\
                              &                                                                           & $R-T-R$ & 573,616    & 0.79 & 0.05 \\
                              &                                                                           & $R-S-R$ & 3,402,743  & 0.77 & 0.05 \\
                              &                                                                           & $ALL$   & 3,846,979  & 0.77 & 0.07 \\ \midrule
\multirow{4}{*}{Amazon}       & \multirow{4}{*}{\begin{tabular}[l]{@{}l@{}}11944\\ (9.5\%)\end{tabular}}  & $U-P-U$ & 175,608    & 0.61 & 0.19 \\
                              &                                                                           & $U-S-U$ & 3,566,479  & 0.64 & 0.04 \\
                              &                                                                           & $U-V-U$ & 1,036,737  & 0.71 & 0.03 \\
                              &                                                                           & $ALL$   & 4,398,392  & 0.65 & 0.05 \\ \midrule
\multirow{4}{*}{ShortMessage} & \multirow{4}{*}{\begin{tabular}[l]{@{}l@{}}16946\\ (8.2\%)\end{tabular}}  & $M-U-M$ & 972,561    & 0.90 & 0.01 \\
                              &                                                                           & $M-T-M$ & 30,698     & 0.47 & 0.12 \\
                              &                                                                           & $M-S-M$ & 35,253,869 & 0.93 & 0.01 \\
                              &                                                                           & $ALL$   & 35,378,957 & 0.93 & 0.01 \\ \bottomrule
\end{tabular*}
\label{tab2}
\end{center}
\end{table*}
% \end{table}

We utilize three real-world datasets (i.e., YelpChi dataset, Amazon dataset, and ShortMessage dataset) to validate the performance of the proposed method. Table\ref{tab2} lists the statistics of all datasets.

The YelpChi dataset is a public dataset that includes hotel and restaurant reviews that are filtered (spam) or recommended (legitimate) by Yelp. We take 32 handcrafted features from \cite{Yelp_data} as original features of reviews for the YelpChi dataset. After preprocessing, the final dataset contains 45,954 reviews (14.5\% spams). Similar to \cite{CARE-GNN}, we conduct a binary classification task for identifying spam reviews in the YelpChi dataset.

The Amazon dataset \cite{Amazon} is a public dataset that includes reviews of products under the musical instruments category. Users in this dataset are labeled legitimate (with more than 80\% helpful votes) and fraudulent (with less than 20\% helpful votes). We take 25 handcrafted features from \cite{AmazonData} as original user features. Similar to previous work \cite{CARE-GNN}, we conduct a binary classification task for identifying fraudulent users in the Amazon dataset.

The ShortMessage dataset includes short messages collected from a specific area of China on August 15 from 0 o’clock to 8 o’clock. We filter messages without sender ID or receiver ID. Then, the short messages are manually labeled as spam and normal by domain experts. After preprocessing, the final dataset contains 16,946 messages (8.2\% spams). In addition, we transform the content of messages into 100-dimensional vectors through Word2Vec \cite{word2Vec} and take them as original features. We conduct a binary classification task for identifying spam messages in the ShortMessage dataset.

\subsection{Graph Construction}
We conduct an undirected multi-relation graph on each dataset. Here are the relations involved:

The YelpChi dataset:
1) $R-U-R$ connects reviews posted by the same user;
2) $R-S-R$ connects reviews under the same product and with the same star rating;
3) $R-T-R$ connects reviews under the same product and posted in the same month.

The Amazon dataset:1) $U-P-U$ connects users reviewing at least one same product;
2) $U-S-U$ connects users having at least one same star rating within one week;
3) $U-V-U$ connects users having the top 5\% mutual review text similarities (measured by TF-IDF) among all users.

The ShortMessage dataset:
1) $M-U-M$ connects messages posted by the same user;
2) $M-T-M$ connects messages posted within a minute;
3) $M-S-M$ connects messages with the same symbols (i.e., phone numbers, URLs).

\subsection{Baseline Methods}

To verify the effectiveness of our proposed method in fraud detection, we compare our model with several state-of-the-art GNN methods and a variant.

\begin{table*}[width=1\textwidth]
\begin{center}
\caption{Performance comparison between HA-GNN and the baselines on the YelpChi, Amazon, and ShortMessage datasets.}
\begin{tabular*}{\hsize}{@{}@{\extracolsep{\fill}}lllllll@{}}
\toprule
\multirow{2}{*}{Model} & \multicolumn{2}{l}{YelpChi}          & \multicolumn{2}{l}{Amazon}         & \multicolumn{2}{l}{ShortMessage}    \\ \cmidrule(l){2-3} \cmidrule(l){4-5} \cmidrule(l){6-7}
                       & AUC              & Recall           & AUC              & Recall           & AUC              & Recall           \\ \midrule
GAT                    & 56.24\%          & 54.52\%          & 75.16\%          & 65.51\%          & 69.98\%          & 68.25\%          \\
GraphSAGE              & 54.00\%          & 52.86\%          & 75.27\%          & 70.16\%          & 72.79\%          & 72.80\%          \\
FdGars                 & 61.75\%          & 58.69\%          & 79.73\%          & 75.16\%          & 76.03\%          & 79.33\%          \\
GEM                    & 64.98\%          & 50.36\%          & 89.04\%          & 84.83\%          & 90.99\%          & 78.75\%          \\
GraphConsis            & 62.07\%          & 62.08\%          & 85.50\%          & 85.53\%          & 76.59\%          & 76.45\%          \\
CARE-GNN               & 75.70\%          & 71.92\%          & 89.73\%          & 88.48\%          & 90.50\%          & 85.16\%          \\ \midrule
HA-GNN-F               & \textit{84.38\%} & \textit{77.31\%} & \textit{90.32\%} & \textit{88.54\%} & \textit{93.03\%} & \textit{85.58\%} \\
HA-GNN                 & \textbf{85.67\%} & \textbf{79.79\%} & \textbf{92.94\%} & \textbf{89.82\%} & \textbf{97.31\%} & \textbf{90.12\%} \\ \bottomrule
\end{tabular*}
% }
\label{tab-result}
\end{center}
\end{table*}

\begin{itemize}
\item GAT \cite{GAT}: It is a general GNN method that learns different weights for different nodes in a neighborhood.
\item GraphSAGE \cite{GraphSAGE}: It is a general GNN method that samples neighboring nodes before aggregation.
\item GEM \cite{GEM}: It is a GNN-based malicious accounts detection that applies an attention mechanism into the aggregation process.
\item FdGars \cite{FdGars}: It is a spam reviewer detection that uses a two-layer GNN to learn node representation.
\item GraphConsis \cite{GraphConsis}: It is a GNN-based fraud detection algorithm that filters inconsistent neighbors according to sampling thresholds.
\item CARE-GNN \cite{CARE-GNN}: It is a GNN-based fraud detection model which enhances the GNN aggregation process with an adaptive neighbor-filter.
\item HA-GNN-F: It is a variant of HA-GNN, which takes the concatenation of local and original node features as the neighborhood attention module input. 
\end{itemize}

\begin{table*}[width=1\textwidth]
% \begin{table}
\begin{center}
\caption{The testing performance for different variants of HA-GNN.}
\begin{tabular*}{\hsize}{@{}@{\extracolsep{\fill}}lllllll@{}}
\toprule
\multirow{2}{*}{Model}             & \multicolumn{2}{l}{YelpChi}          & \multicolumn{2}{l}{Amazon}         & \multicolumn{2}{l}{ShortMessage}    \\ \cmidrule(l){2-3} \cmidrule(l){4-5} \cmidrule(l){6-7} 
      & AUC              & Recall           & AUC              & Recall           & AUC              & Recall           \\ \midrule
HA-GNN$\_1$ & 77.85\%          & 69.54\%          & 84.04\%          & 69.76\%          & 91.39\%          & 80.93\%          \\
HA-GNN$\_2$ & 80.89\%          & 74.89\%          & 84.22\%          & 76.67\%          & 92.94\%          & 83.36\%          \\
HA-GNN      & \textbf{85.67\%} & \textbf{79.79\%} & \textbf{92.94\%} & \textbf{89.82\%} & \textbf{97.31\%} & \textbf{90.12\%} \\ \bottomrule
\end{tabular*}
\label{tab-ablation}
\end{center}
% \end{table}
\end{table*}

\subsection{Experimental Setup} \label{experimental setup}
For HA-GNN, the dimension of final node embedding is set as 32. The number of attention heads $K$ in Eq.\ref{eq_z_v_k} is set as 8. For the YelpChi dataset and the Amazon dataset, the number of epochs is set as 15. For the ShortMessage dataset, the number of epochs is set as 25. For all datasets, we use Adam \cite{Adam} with a learning rate of $5\times 10^{-3}$ for training. For the baselines, we use the open-source implementation \footnote{https://github.com/safe-graph/DGFraud}. The data format is transformed appropriately to fit their settings. All models run on Python 3.7.10, GCC 7.3.0, and 2.10 GHz Intel Core i5 Linux desktops.

\subsection{Evaluation Metrics}
We adopt Area Under ROC Curve (AUC) and Recall as the evaluation metrics, which are broadly used in fraud detection \cite{GEM,CARE-GNN}. It is worth mentioning that Recall is the most concerning metric in practice. The AUC represents the ability of a classifier to rank positive samples in front of negative instances, which avoids the influence of unbalanced data.

\subsection{Camouflage Analysis}
 Similar to \cite{CARE-GNN}, we adopt two characteristic scores to analyze fraudster camouflage. For relation camouflage, we measure it by using the average label similarity between neighboring pairs under each relation $r$:
\begin{equation}
    Avg.Label Similarity = \sum_{\left(u,v\right)\in\mathcal{E}_r}(1 - I(u \sim v)) / |\mathcal{E}_r| \label{eq_s_r}
\end{equation}
where $I\left(\cdot\right)\in\{0, 1\}$ is an indicator function to indicate whether node $u$ and node $v$ are in the same class. We calculate the summation of all the indication w.r.t. all the edges and the results are normalized by the total number of edges $|\mathcal{E}_r|$. The results are presented in Table \ref{tab2}. We observe that only $R-U-R$ relation in the YelpChi dataset has an average label similarity score more than $80\%$, while other relations have low average label similarity scores. It implies that fraudsters camouflage themselves successfully under single relation. The adjacency matrices across multiple relations represent more complex interactions \cite{netgraphs}, which againsts relation camouflage on single relation, thus helping to improve the performance of fraud detection.

For feature camouflage, we calculate the average original feature similarity between neighboring pairs under each relation $r$:
\begin{equation}
 Avg.Feature Similarity = \sum_{\left(u,v\right)\in\mathcal{E}_r} \exp{\left(-\|x_u-x_v\|_2^2\right)}/|\mathcal{E}_r|\cdot d \label{eq_s_f}
\end{equation}
We employ the original feature vectors' Euclidean distance as the similarity measurement between two neighboring nodes. The overall feature similarity score is normalized by the total number of edges $|\mathcal{E}_r|$ and the feature dimension $d$. The results are presented in Table \ref{tab2}. A high average feature similarity score implies fraudsters camouflage their original feature to disguise themselves like legitimate users. We observe that most relations have average label similarity scores more than $60\%$. It indicates that original features may mislead the classification of fraudsters. We need to introduce additional information (such as tie strength) to help distinguish fraudsters.

\subsection{Overall Evaluation}
We conduct experiments on the Yelp, Amazon, and ShortMessage datasets based on the above experimental settings. The results are shown in Table \ref{tab-result}. From the results, we have the following observations:

1) HA-GNN outperforms all the baseline models. It demonstrates the practical design of the proposed model. Traditional GNN methods (i.e., GAT, GraphSAGE, FdGars) detect fraudsters based on single relation. Compared to these methods, HA-GNN incorporates multiple relations through an attention mechanism, thus capturing richer high-order structural information. Filter-based fraud detectors (i.e., GraphConsis and CERA-GNN) filtering neighbors before aggregations, probably be misled by camouflaged original features. Compared to these methods, HA-GNN adaptively allocates different weights to structural and original features, facilitating the identification of camouflaged fraudsters.

\begin{figure*}
\centering
\subfigure[Relation weights on the YelpChi dataset]{
\includegraphics[width=0.47\hsize]{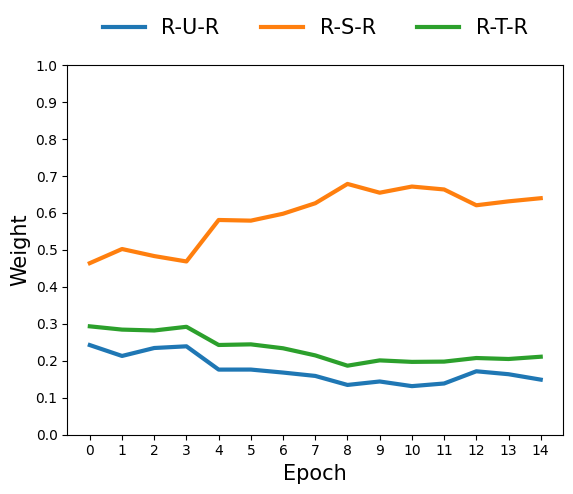}
}
\subfigure[Information weights on the YelpChi dataset]{
\includegraphics[width=0.47\hsize]{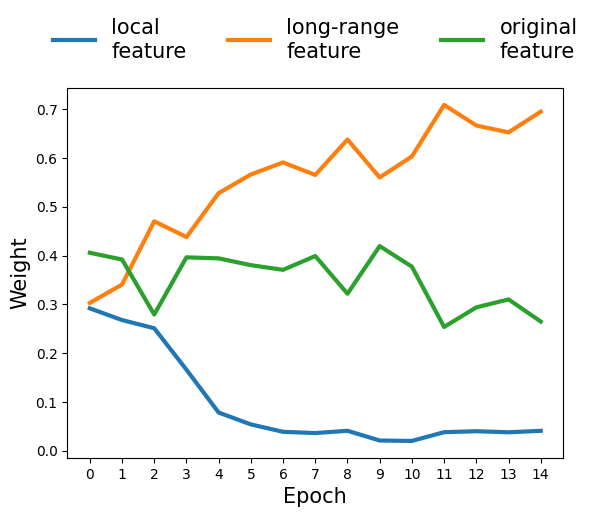}}
\caption{\centering{The training process of the proposed model on the YelpChi dataset.}}
\label{fig-analysis-YelpChi-attention}
\end{figure*}

\begin{figure*}
\centering
\subfigure[Relation weights on the Amazon dataset]{
\includegraphics[width=0.47\hsize]{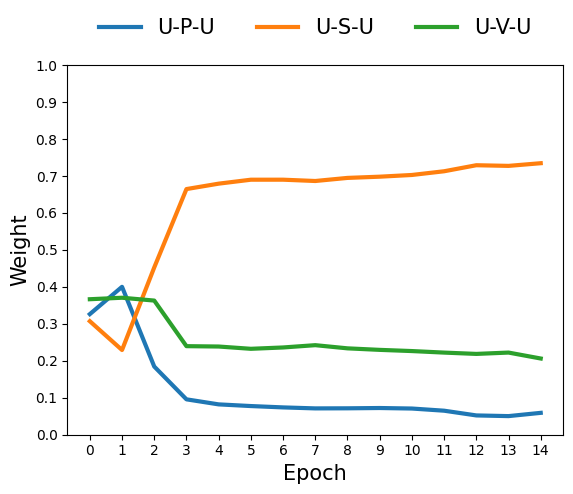}
}
\subfigure[Information weights on the Amazon dataset]{
\includegraphics[width=0.47\hsize]{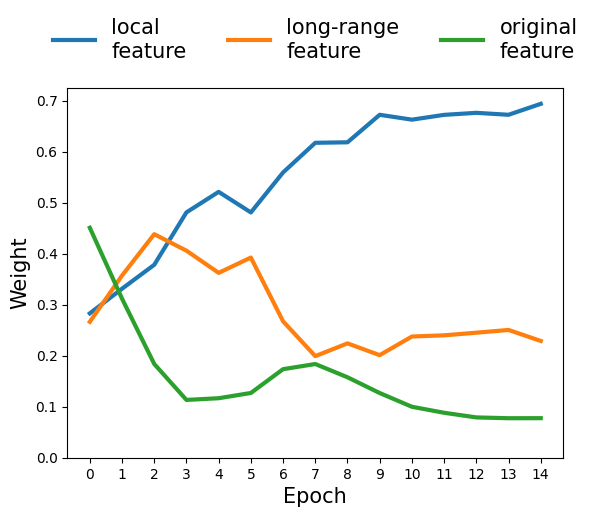}}
\caption{\centering{The training process of the proposed model on the Amazon dataset.}}
\label{fig-analysis-Amazon-attention}
\end{figure*}

2) HA-GNN-F outperforms all the baseline models: The performance of HA-GNN-F is better than those of all the other baselines, which demonstrates the effect of the hierarchical attention mechanism. Compared to other GNN-based models learning node representation through aggregating original neighbor features, HA-GNN-F exploits local and long-range structures, therefore better handling fraudsters with relation and feature camouflage.

3) HA-GNN is consistently superior to HA-GNN-F: HA-GNN achieves the best performance. It indicates that aggregating original node features of neighbors have a negative effect on learning discriminate node representations for fraud detection. HA-GNN-F aggregates neighbors’ local and original features, which smooth out the suspicion of fraudsters with legitimate neighbors. Instead of involving original neighbor features, HA-GNN is focused on neighbors’ local features, which helps improve the performance much more.

\begin{figure*}
\centering
\subfigure[Relation weights on the ShortMessage dataset]{
\includegraphics[width=0.47\hsize]{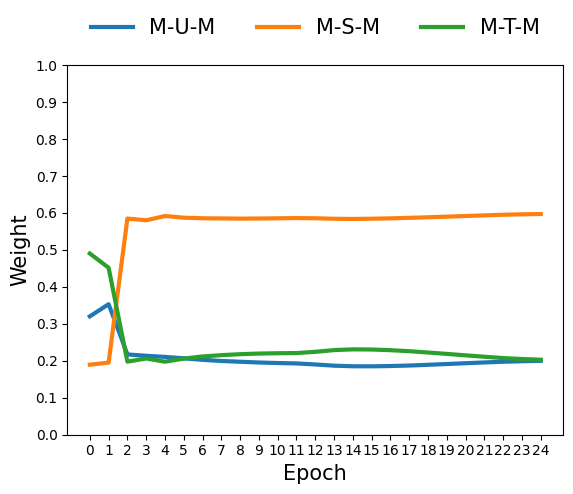}
}
\subfigure[Information weights on the  ShortMessage dataset]{
\includegraphics[width=0.47\hsize]{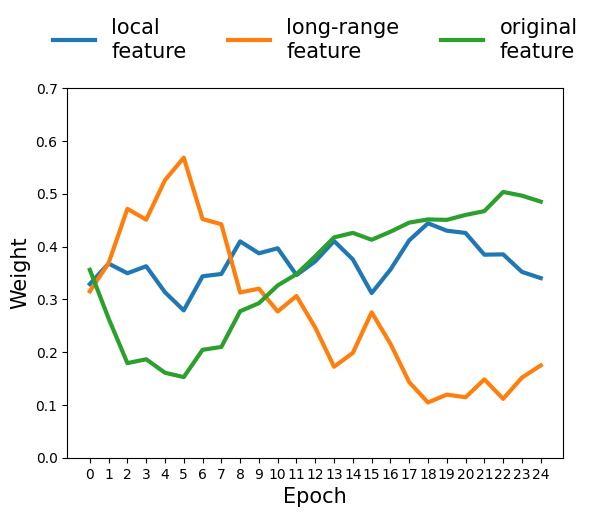}
}
\caption{\centering{The training process of the proposed model on the ShortMessage dataset.}}
\label{fig-analysis-ShortMessage-attention}
\end{figure*}

\subsection{Ablation Study}
To study the effectiveness of the three modules, we perform two variant experiments to show that all these modules contribute to fraud detection. The results of AUC and Recall are presented in Table \ref{tab-ablation}. HA-GNN$\_1$ denotes the variant with only the relation attention module, HA-GNN$\_2$ denotes the variant with both the relation attention module and neighborhood attention module. The corresponding observations are two-fold: (1) variants with combined modules perform better. For example, the variant with the relation attention module and neighborhood attention module outperforms variant with only the relation attention module. The main reason is that variants with combined modules can extract more comprehensive information of nodes and learn highly expressive node representations. (2) the complete model HA-GNN outperforms all other variants. It verifies the effectiveness of the proposed information fusion module.

\subsection{Aggregation Process Analysis}
Figure \ref{fig-analysis-YelpChi-attention} (a), Figure \ref{fig-analysis-Amazon-attention} (a), and Figure \ref{fig-analysis-ShortMessage-attention} (a) show the change of relation weights on the YelpChi dataset, the Amazon dataset, and the ShortMessage dataset, respectively. The relation weights averaged over all nodes in the graph for a fair comparison. As the training epoch increases, the weights of relations are updated and quickly converge to steady states. It implies that the relation attention module can quickly determine the importance of different relations within the first few epochs and lead the following training processes. 

To illustrate the different importance of different features on complex real-world datasets, we plot the weight trends of three kinds of features during the training process. The weights averaged over all nodes in the graph for a fair comparison. The results on the YelpChi dataset, the Amazon dataset, and the ShortMessage dataset are shown in Figure \ref{fig-analysis-YelpChi-attention} (b), Figure \ref{fig-analysis-Amazon-attention} (b), and Figure \ref{fig-analysis-ShortMessage-attention} (b), respectively. With training processing, the weights of local features, long-range features, and original node features converge to different values on different datasets. For example, in the late stages of training, the weight of the long-range feature is the highest on the YelpChi dataset, while the weight of the original node feature is the highest on the ShortMessage dataset. It indicates the ability of the proposed information fusion module to adaptively find the most important feature for distinguishing fraudsters on different datasets.

\begin{figure}
\centerline{\includegraphics[width=0.47\textwidth]{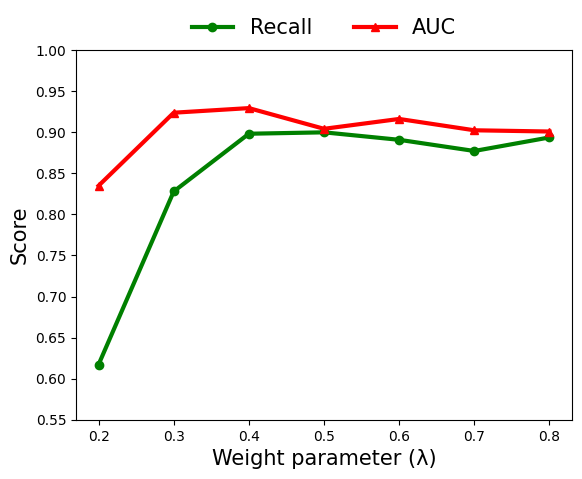}}
\caption{\centering{Impact of the weight parameter $\lambda$.}}
\label{fig-Parameter Sensitivity}
\end{figure}

\subsection{Hyper-parameter Sensitivity}
This section investigates the sensitivity of an important hyper-parameter. Taking the Amazon dataset as an example, we study the influence of the weight parameter $\lambda$ from 0.2 to 0.8. The rest of the parameters are set the same as in Section \ref{experimental setup}. Experimental results are presented in Figure \ref{fig-Parameter Sensitivity}. From Figure \ref{fig-Parameter Sensitivity}, we observe that the performance curve first raises and then slightly drops, and HA-GNN achieves the best performance when the value of $\lambda$ is set to 0.4. When $\lambda$ changes from 0.2 to 0.4, the performance becomes better, which indicates that paying too much attention to legitimate instances has little effect on fraud detection and will bring noise, thus reducing the detection performance. When $\lambda$ is greater than 0.4, the performance becomes worse, then finally tends to be stable. It indicates that incorporating weight parameter $\lambda$ to balance the classes is of great significance for fraud detection. However, experimental results are also acceptable when $\lambda$ is greater than 0.4, indicating the proposed method's stability.

\section{Conclusion}
In this paper, we developed a novel hierarchical attention network model, named HA-GNN that improves node embeddings for fraud detection with camouflage. Specifically, HA-GNN explored a relation attention module to aggregate the adjacency matrices across different relations, and obtained a local embedding for each node. To grasp the structural information of long-range scope, HA-GNN further devised a multi-head self-attention mechanism as the neighborhood attention module to learn a long-range embedding for each node. HA-GNN generated the final node embeddings by integrating the structural information into original node features. Our empirical studies showed that HA-GNN achieved better detection performance than the state-of-the-arts, and verified the benefits of the proposed hierarchical attention mechanism.

\bibliographystyle{cas-model2-names}
% \bibliographystyle{unsrt}

\bibliography{cas-refs}

\end{document}